\documentclass[runningheads]{llncs}
\usepackage{graphicx}
\usepackage{marvosym}
\usepackage{amsmath}
\usepackage{subfigure} 
\usepackage{float}
\usepackage{stfloats}
\usepackage{booktabs}
\usepackage{setspace}
\usepackage{amssymb}
\usepackage{multirow}
\usepackage{ifsym}
\usepackage{cite}
\usepackage[colorlinks, bookmarks=false,linkcolor=blue, citecolor=blue,urlcolor=blue]{hyperref}
\begin{document}
\title{FIT: Frequency-based Image Translation for Domain Adaptive Object Detection\thanks{This work was supported in part by the National Key Research and Development Plan of China under Grant 2020AAA0108902 and the Strategic Priority Research Program of Chinese Academy of Science under Grant XDB32050100.}}
\titlerunning{Frequency-based Image Translation for Domain Adaptive Object Detection}
%
\author{Siqi~Zhang\inst{1,2} \and
Lu~Zhang\inst{1}\and
Zhiyong~Liu\inst{1,2\textrm{(\Letter)}}\and
Hangtao~Feng\inst{1,2}
}
\authorrunning{S. Zhang et al.}
%
\institute{State Key Laboratory of Management and Control for Complex Systems,\\
Institute of Automation, Chinese Academy of Sciences,\\
Beijing 100190, China\\
\email{\{zhangsiqi2020, lu.zhang, zhiyong.liu, fenghangtao2018\}@ia.ac.cn}\\
\and
School of Artificial Intelligence, University of Chinese Academy of Sciences, \\Beijing 100190, China}
\toctitle {FIT: Frequency-based Image Translation for Domain Adaptive Object Detection}
\tocauthor{Siqi~Zhang, Lu~Zhang, Zhiyong~Liu, Hangtao~Feng}
\maketitle              
\begin{abstract}
Domain adaptive object detection (DAOD) aims to adapt the detector from a labelled source domain to an unlabelled target domain. In recent years, DAOD has attracted massive attention since it can alleviate performance degradation due to the large shift of data distributions in the wild. To align distributions between domains, adversarial learning is widely used in existing DAOD methods. However, the decision boundary for the adversarial domain discriminator may be inaccurate, causing the model biased towards the source domain. To alleviate this bias, we propose a novel Frequency-based Image Translation (FIT) framework for DAOD. First, by keeping domain-invariant frequency components and swapping domain-specific ones, we conduct image translation to reduce domain shift at the input level. Second, hierarchical adversarial feature learning is utilized to further mitigate the domain gap at the feature level. Finally, we design a joint loss to train the entire network in an end-to-end manner without extra training to obtain translated images. Extensive experiments on three challenging DAOD benchmarks demonstrate the effectiveness of our method.
\end{abstract}
\section{Introduction}
In recent years, object detectors~\cite{ren2015faster,redmon2016you,tian2019fcos} based on deep convolutional neural networks have demonstrated outstanding performance on a variety of datasets. However, existing object detection models still face serious challenges when deployed in practice such as autonomous driving and robotic manipulation, due to various changes in weather, illumination, object appearance, \textit{etc}. \begin{figure}[ht]
\begin{minipage}{0.32\linewidth}
\centerline{\includegraphics[width=\textwidth]{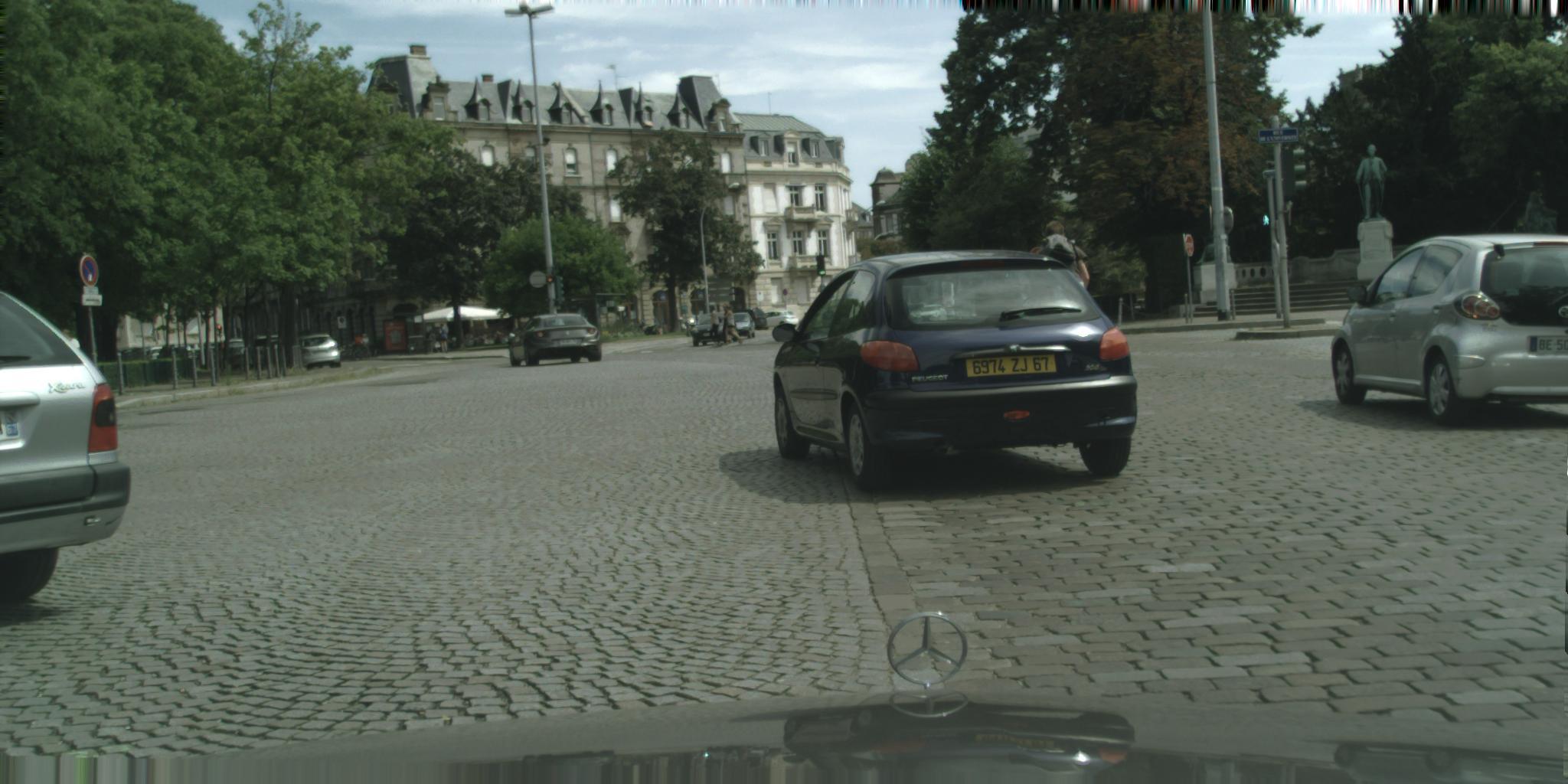}}
\end{minipage}
\begin{minipage}{0.32\linewidth}
\centerline{\includegraphics[width=\textwidth]{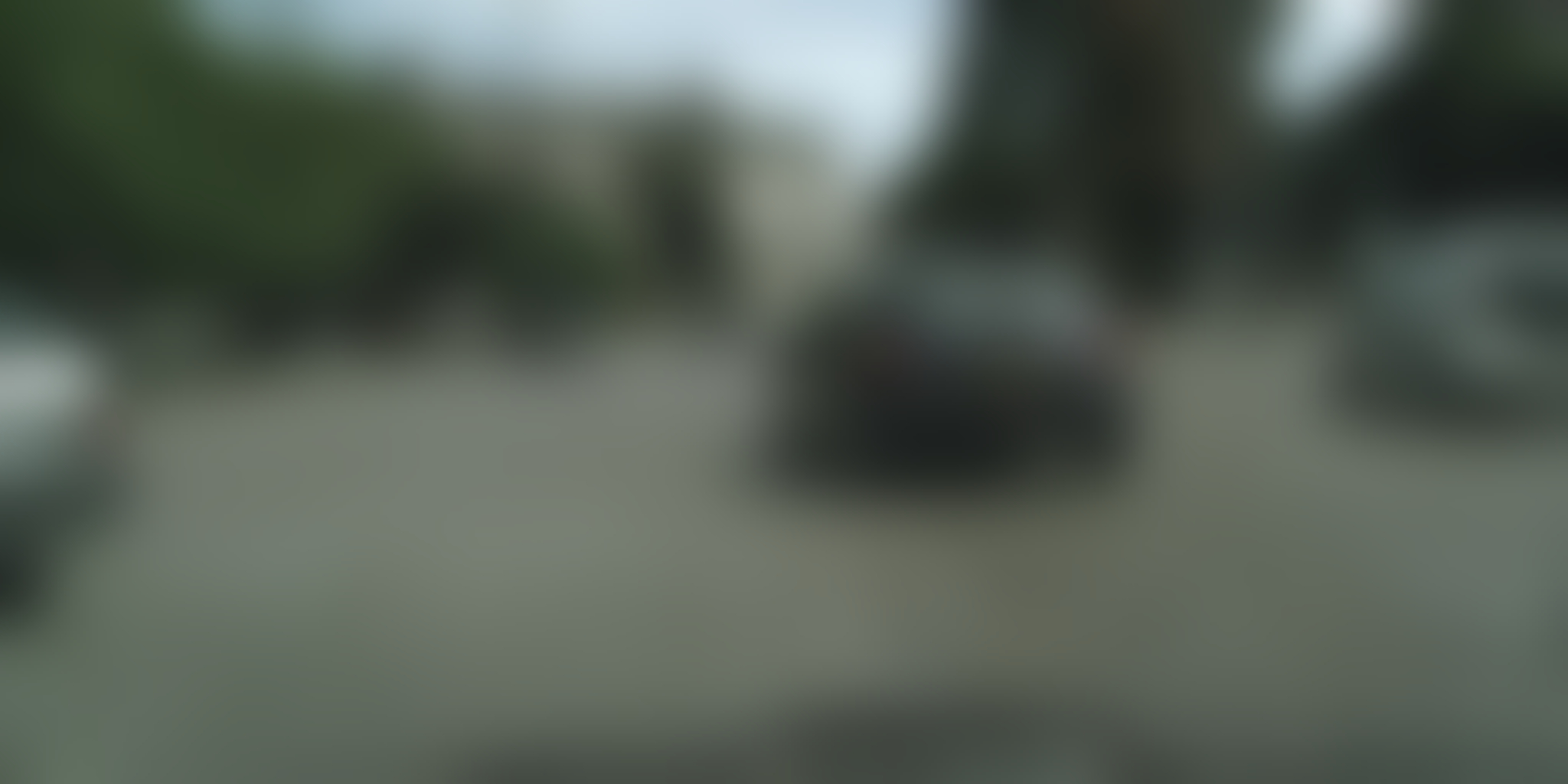}}
\end{minipage}
\begin{minipage}{0.32\linewidth}
\centerline{\includegraphics[width=\textwidth]{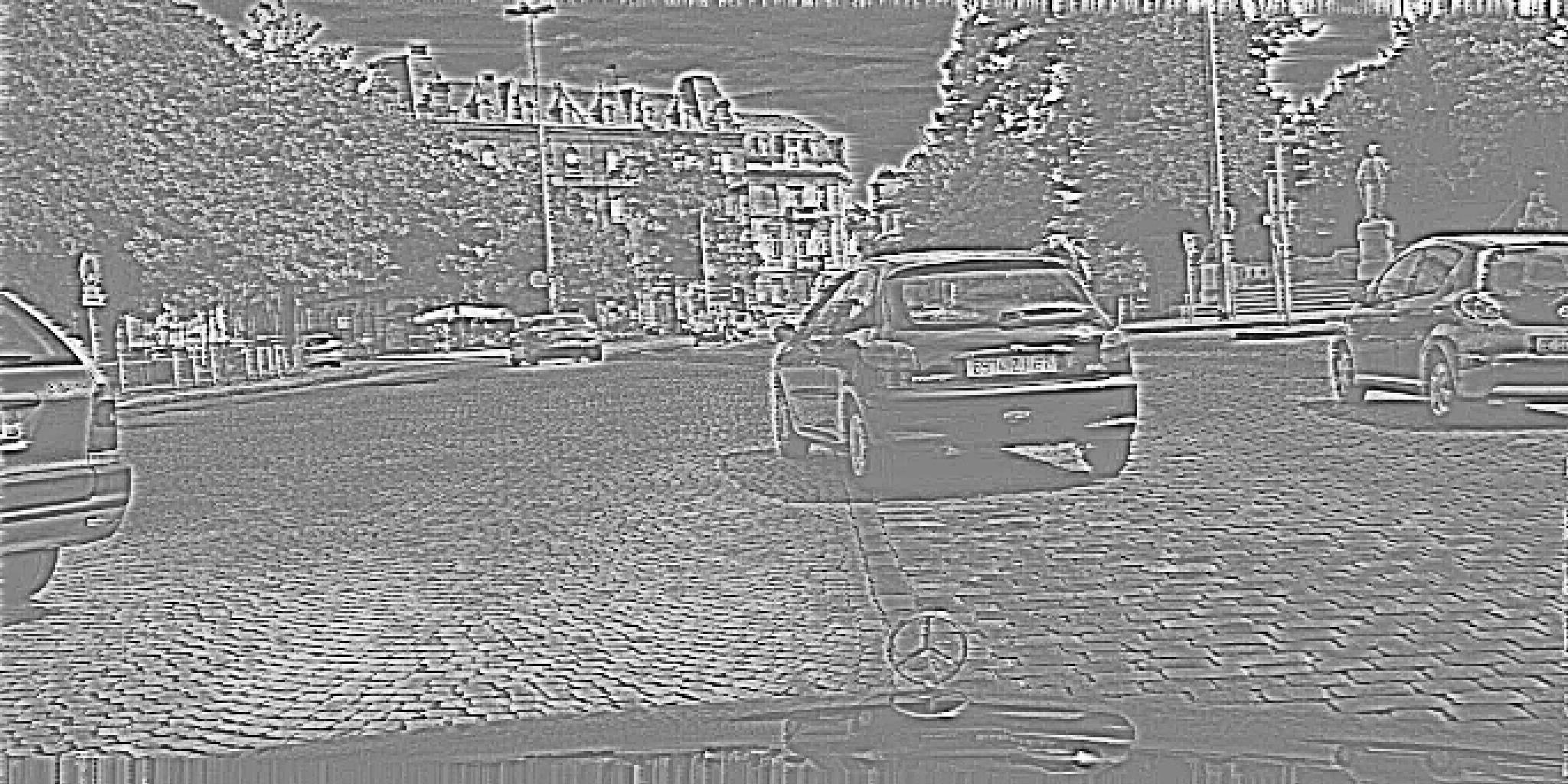}}
\end{minipage}\\
\begin{minipage}{0.32\linewidth}
\centerline{\includegraphics[width=\textwidth]{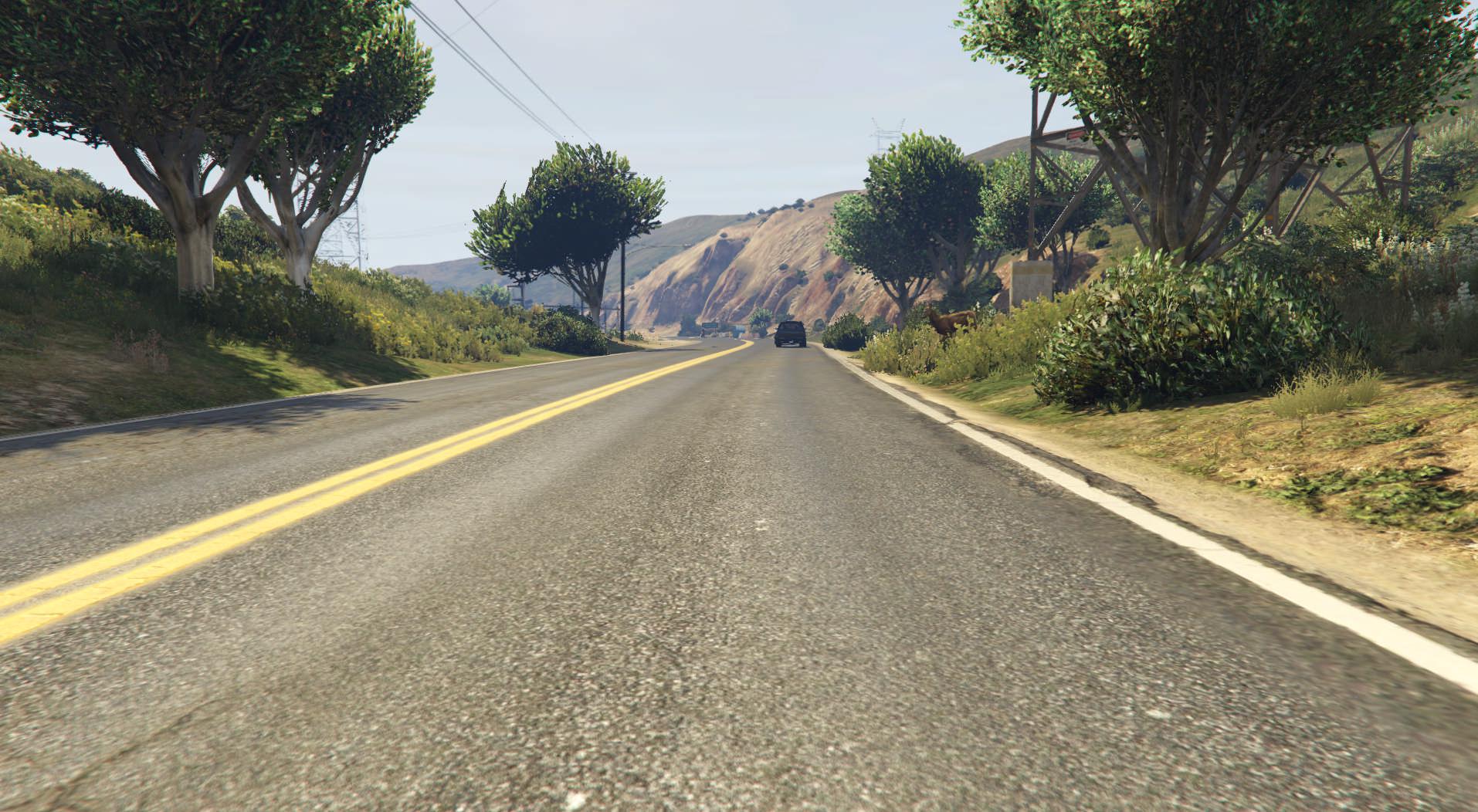}}
\centerline{(a)}
\end{minipage}
\begin{minipage}{0.32\linewidth}
\centerline{\includegraphics[width=\textwidth]{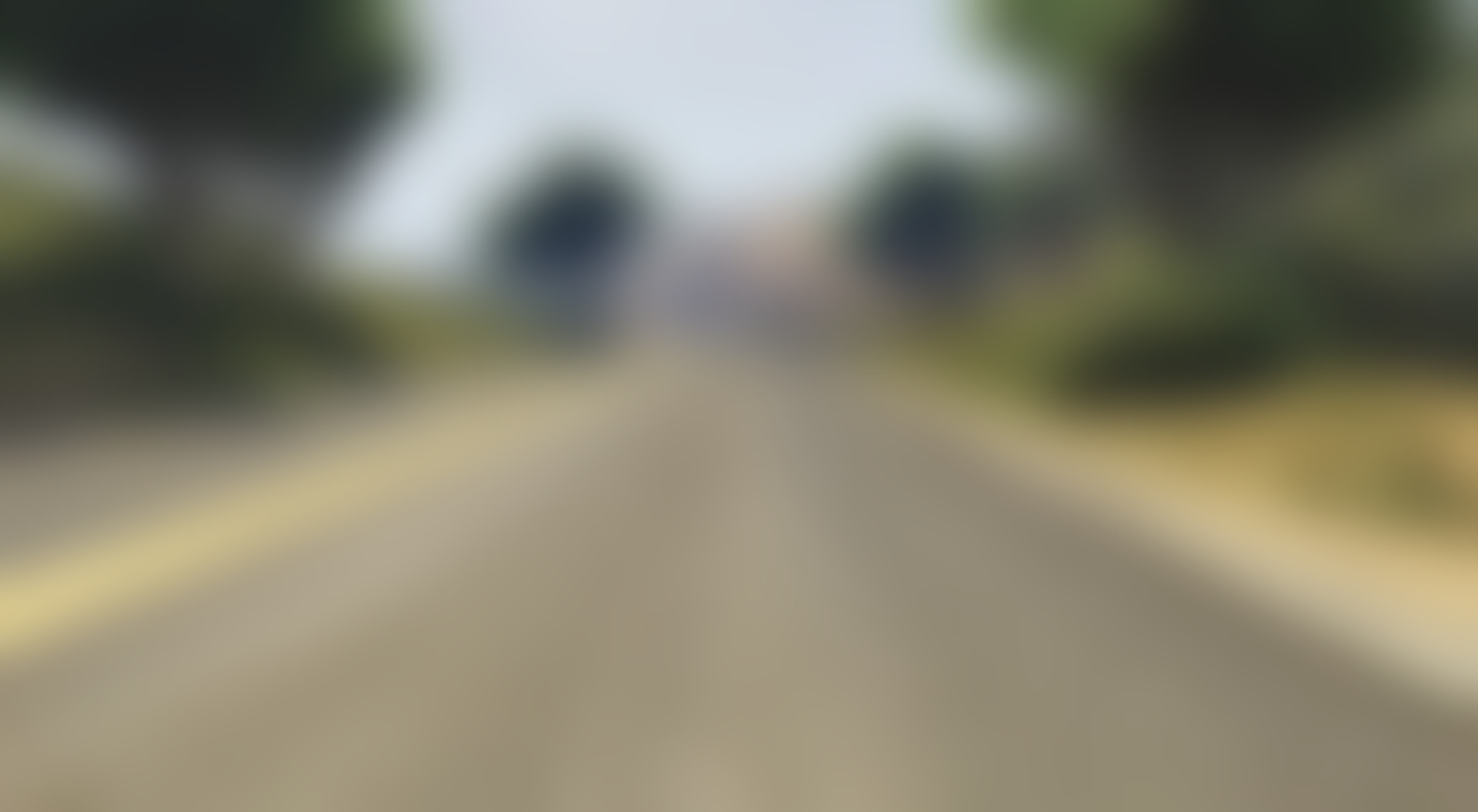}}
\centerline{(b)}
\end{minipage}
\begin{minipage}{0.32\linewidth}
\centerline{\includegraphics[width=\textwidth]{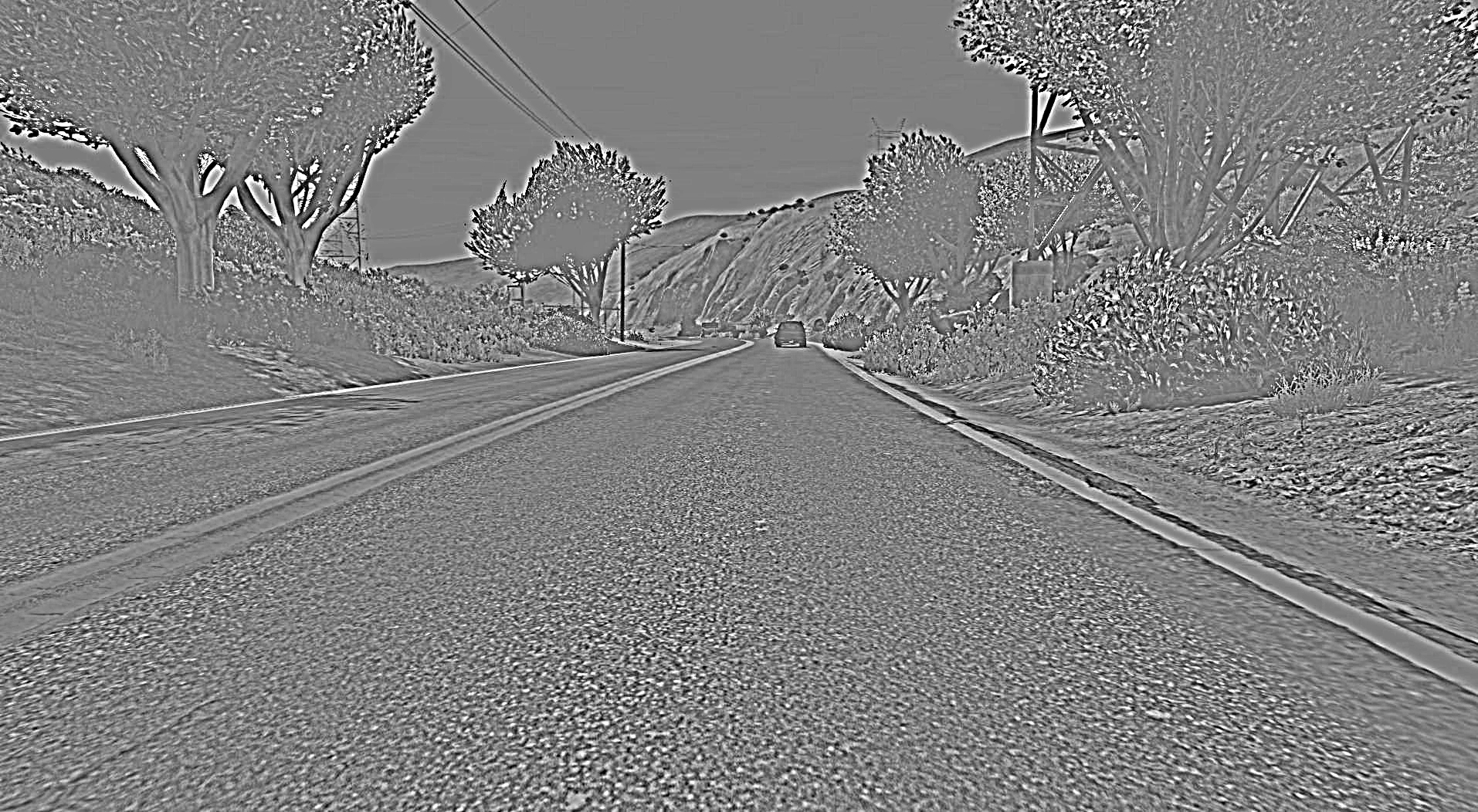}}
\centerline{(c)}
\end{minipage}
\caption{Visualization of frequency decomposition of source image: (a), (b) and (c) show original image, low-pass and high-pass filtered image.}
\label{fig1}
\end{figure}These changes may lead to domain gaps between the training and testing data, which has been observed to cause dramatic drops in the performance of the trained detector~\cite{chen2018domain}.
Although we can annotate for each new data to mitigate the problem, it is costly and even infeasible because of the countless situations in the real world. Therefore, adaptive object detectors that can bridge the domain gap from the source to the target domain are highly desirable.

Domain adaptive object detection (DAOD), which trains with labelled source datasets and unlabelled target datasets, aims to tackle domain shift to get better performance on the visually distinct target domain. Many previous works~\cite{chen2018domain,saito2019strong,su2020adapting,he2020domain,wu2021vector,zhang2021rpn,liu2022decompose} attempt to utilize adversarial feature learning~\cite{ganin2015unsupervised} to align feature distributions to extract domain-invariant features. But the adversarial training process could be unstable~\cite{wu2021vector,liu2022decompose}, which makes the decision boundary for the adversarial domain discriminator inaccurate, causing the model biased towards the source domain. To alleviate this problem, some methods~\cite{kim2019diversify,Chen_2020_CVPR,hsu2020progressive,shen2021cdtd} utilize the image translation model GANs, like CycleGAN~\cite{zhu2017unpaired} to translate source images to target-like images or vice versa to further mitigate the domain gap and make the detector perform better on the target datasets. However, GANs for domain adaption object detection have two following
limitations. First, GANs could fail to keep semantic consistency and tend to lose important structural characteristics~\cite{chen2021ssd}. Second, GANs-based methods need extra training to prepare translated images before training the adaptive detector, which is time-consuming.

To address the above limitations, we propose a novel Frequency-based Image Translation method to mitigate the input-level domain gap without extra time-consuming training. Inspired by digital signal processing theories \cite{oppenheim1999discrete}, we exploit the frequency information to translate the image style and maintain semantic consistency. Intuitively, the low-frequency component largely captures domain-specific information, such as colours and illuminations\cite{piotrowski1982demonstration}, while the high-frequency component mainly obtains domain-invariant information, such as edges and shapes, which are important details of objects \cite{li2015finding}, as shown in Fig.\ref{fig1}. Motivated by this, we present the Frequency-based Image Translation (FIT) module, which decomposes the image into multiple frequency components, keeps domain-invariant frequency components unchanged and swaps domain-specific ones. Moreover, a novel module called Frequency Mask is designed to identify whether the frequency component is domain-specific in FIT. Then, hierarchical adversarial feature learning is utilized to further boost the performance. The entire network can be optimized in an end-to-end manner under the supervision of a joint loss function.
\begin{figure*}[t]
\centering
\includegraphics[width=\textwidth]{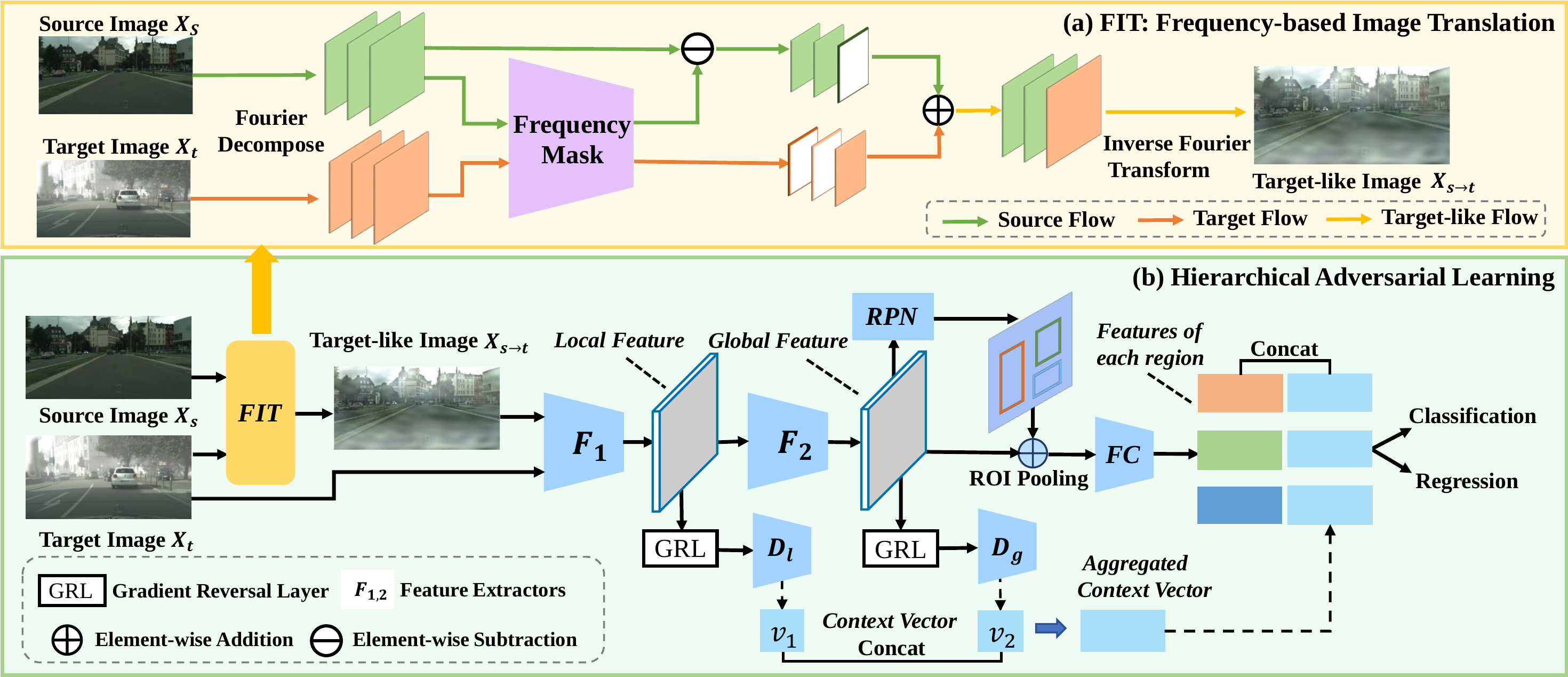}
\caption{Overview of the proposed framework. (a) illustrates Frequency-based Image Translation (FIT) module, where green arrows represent the flow of source data and orange arrows represent the flow of target data. The overall pipeline is illustrated in (b). The source $X_s$ and target $X_t$ images are fed into FIT to obtain target-like images $X_{s\rightarrow t}$, and $X_{s\rightarrow t}$ and $X_t$  are as the inputs for the object detector. We align the local and global feature by a local domain classifier $D_l$ and a global domain classifier $D_g$. $F_1$ and $F_2$ denote the different level feature extractors. The context vectors $v_{1,2}$ are extracted by the domain classifiers and concatenated with features of regions before the final fully connected layer.
}
\label{fig2}
\end{figure*}
The contributions of this work can be summarized as follows: 
  \begin{itemize}
    \item A novel Frequency-based Image Translation (FIT) method is presented for DAOD, which leverages frequency information to mitigate the domain shift at the input level. To further boost the adaptation performance, we introduce hierarchical adversarial learning to align distributions at the feature level.
    \item Different from traditional GANs-based methods, the entire network can be trained in an end-to-end manner without extra time-consuming training, since the proposed frequency-based image translation is embedded as a module in the detection network.
    \item We conduct extensive experiments on three challenging DAOD benchmarks and our FIT achieves favorable performance under various domain-shift scenarios, demonstrating the effectiveness of the proposed method.
\end{itemize} 
\section{Proposed Method}
\subsection{Overview}
\textbf{Problem Definition. }The domain adaptation~\cite{ganin2015unsupervised} task typically considers two domains: the source domain $S$ and target domain $T$. Specifically, we have access to a labelled source dataset $\mathcal{D}_{s}=\left\{\left(x_{i}^{s}, y_{i}^{s}\right)\right\}_{i=1}^{n_{s}}$ and a target dataset $\mathcal{D}_{t}=\left\{x_{j}^{t}\right\}_{j=1}^{n_{t}}$ with no ground-truth annotations. Here, $x_{i}^{s}$ denotes $i^{th}$ source image, $y_{i}^{s}$ denotes the corresponding label and ${n_{s}}$ denotes the number of source images. Similarly, $x_{j}^{t}$ denotes $j^{th}$ target image and ${n_{t}}$ denotes the number of target images. The source and target domains with different data distributions share the same label space, and the goal of domain adaptive object detection is to train an detector with $\mathcal{D}_{s}$ and $\mathcal{D}_{t}$, which performs well on the target dataset $\mathcal{D}_{t}$. Following the mainstream domain adaptive object detection methods~\cite{chen2018domain,saito2019strong,su2020adapting,he2020domain,wu2021vector,zhang2021rpn,liu2022decompose}, the proposed method is based on the Faster RCNN~\cite{ren2015faster} framework.

\noindent
\textbf{Overall framework. }The overall framework of the proposed method is shown in Fig. \ref{fig2}. We first transform source images $X_s$ to target-like images $X_{s\rightarrow t}$ via frequency-based image translation (FIT), as shown in Fig.\ref{fig2}(a). The key idea is to decompose the image into multiple frequency components and then feed them to the Frequency Mask to identify domain-specific frequency components. Then we replace the domain-specific components of the source image with the corresponding ones of the target image and get the target-like image $X_{s\rightarrow t}$ via the Inverse Fourier Transform. Afterwards, we put target-like images $X_{s\rightarrow t}$ and target images $X_{t}$ into object detector and align the local and global feature by hierarchical adversarial learning, as shown in Fig.\ref{fig2}(b). Through this framework, the domain gap at both input and feature level can be mitigated. The details of the proposed method are given in the following sections.
\subsection{Frequency-based Image Translation}
 In order to mitigate the domain gap at the input level, a novel frequency-based image translation is presented to obtain translated images without changing their semantic structures. The framework of frequency-based image translation is shown in Fig.\ref{fig2}(a). 
 
 First, Fourier transform $\mathcal{F}(\cdot)$ is performed on the image $x$ of size $H \times W$:
 \begin{equation}
\mathcal{F}(x)(a,b)=\sum_{h=0}^{H-1} \sum_{w=0}^{W-1} x(h, w)e^{ -i2 \pi  \cdot\left( \frac{h a}{H}+\frac{w b}{W}\right)}, \label{3}
\end{equation}
for $a=0, \ldots, H-1,$ $b=0, \ldots, W-1$.

Then, we decompose the frequency space representation $\mathcal{F}(x)$ of the image into $N$ components $\left\{x  ^{1}, x^{2}, \ldots, x^{N}\right\}$ of equal bandwidth via band-pass filter $\mathcal{B}(\cdot ; \cdot )$: 
\begin{equation}
x^{fs}=\mathcal{B}(\mathcal{F}(x) ; N)=\left\{x  ^{1}, x^{2}, \ldots, x^{N-1}, x^{N}\right\}, \label{1}
\end{equation} 
\begin{equation}
x^{n} = \begin{cases}\mathcal{F}(x)(i, j), & \text { if } \frac{n-1}{N}<d\left((i, j),\left(c_{i}, c_{j  }\right)\right)<\frac{n}{N} \\ 0
, &\text { otherwise }\end{cases}, \label{2}
\end{equation}
where $c_i$ and $c_j$ denote the image centroid, $d(\cdot, \cdot  )$ denotes the Euclidean distance, and $N$ is the number of components. In our experiments, we set $N=64$.
\begin{figure*}[t]
\centering
\includegraphics[width=0.95\textwidth]{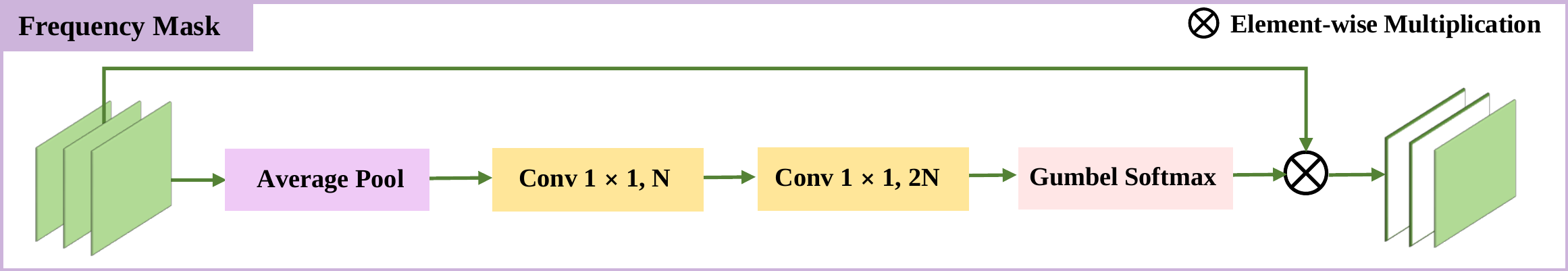}
\caption{ Structure of Frequency Mask.}
\label{fig3}
\end{figure*}

To identify which frequency component is domain-specific, we design a module called Frequency Mask and its structure is shown in Fig.\ref{fig3}. Motivated by the Squeeze-and-Excitation Networks~\cite{Hu_2018_CVPR}, which model the interdependencies between the channels and recalibrate the channel-wise feature responses adaptively, we design similar structure and add Gumbel-Softmax~\cite{jang2016categorical} to make the value close to one-hot vector. '1' means the frequency component is domain-specific, while '0' denotes it is domain-invariant. By Frequency Mask, we find the domain-specific components $DS(x_{s}^{fs})$:
\begin{equation}
DS(x_{s}^{fs})=M\left(x_{s}^{fs}\right) \cdot x_{s}^{fs}, \label{add}
\end{equation} 
where $M(x_{s}^{fs})$ represents the output of Gumbel-Softmax in Fig.\ref{fig3}. Then, we replace the domain-specific components of the source image with the corresponding ones of the target image: 
\begin{equation}
\hat{x}_{s \rightarrow t  }^{fs}= x_{s}^{fs} - DS(x_{s}^{fs}) + DS(x_{t}^{fs}), \label{4}
\end{equation} 

After replacing components, we combine all frequency components and perform Inverse Fourier transform $\mathcal{F}^{-1}(\cdot)$. Finally, we obtain the target-like image:
\begin{equation}
 x_{s\rightarrow t}=\mathcal{F}^{-1}\left(\sum \hat{x}_{s\rightarrow t}^{fs}\right). \label{5}
 \end{equation}

In order to keep the consistency of semantic information, we regulate the reconstruction loss: 
 \begin{equation}
\mathcal{L}_{rec}(X)=\left \| H(X)-H(\hat{X})  \right \|_{1}, \label{6}
 \end{equation}
where $X$ and $\hat{X}$ represent the original and translated image. $H(\cdot)$ represents the band-pass filter that extracts the middle and high-frequency components, which largely capture the semantic information.
\subsection{Hierarchical Adversarial Feature Learning}
After the frequency-based image translation, we put target-like and target images into the object detector and further mitigate the feature-level domain gap by the domain classifier and gradient reversal layer (GRL)~\cite{ganin2015unsupervised}. Since different domains could have distinct scene layouts, fully matching the entire distributions of source and target images at the global image-level may fail~\cite{saito2019strong,he2020domain}. Therefore, we adopt different strategies on the local and global features.

\begin{figure}[t]
\centering
\subfigure[Global Domain Classifier $D_g$]{
\begin{minipage}[t]{0.5\linewidth}
\centering
\includegraphics[width=\linewidth]{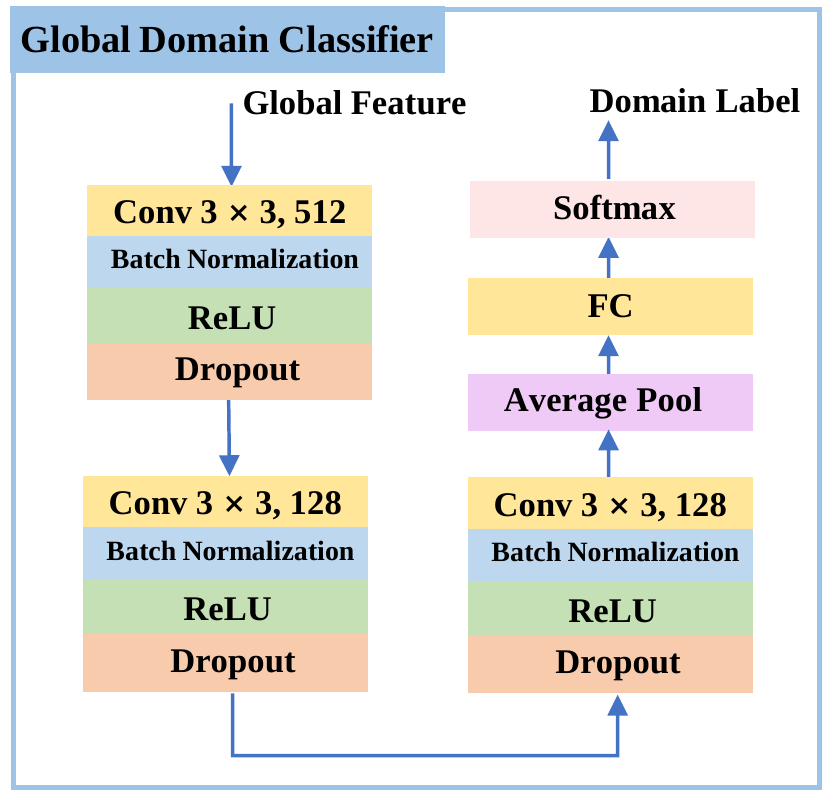}
\end{minipage}%
}%
\subfigure[Local Domain Classifier $D_l$]{
\begin{minipage}[t]{0.425\linewidth}
\centering
\includegraphics[width=\linewidth]{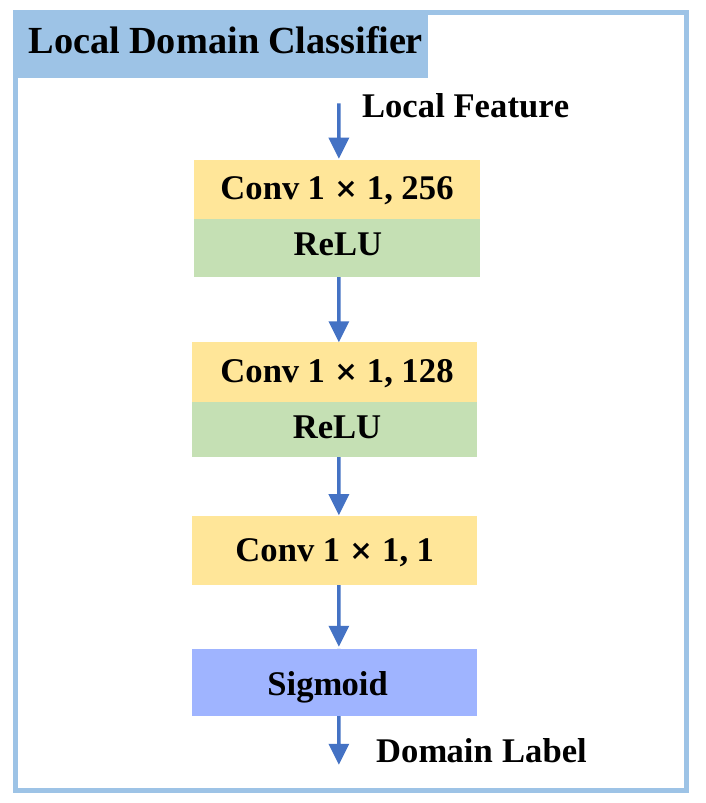}
\end{minipage}
}%
\centering
\caption{Structure of Domain Classifiers.}
\label{fig:d}
\end{figure}

The global feature alignment module consists of a global domain classifier $D_g$ and a GRL. The GRL connects the global domain classifier and the backbone, which reverses the gradients that flow through the backbone, as shown in Fig.\ref{fig2}(b). It means that the global domain classifier $D_g$ aims to distinguish which domain the global feature comes from, whereas the backbone attempts to confuse the classifier. Here, the source images are given the domain label $d = 0$ and the label is 1 for the target images. The loss of the global feature alignment module is calculated as follows,
\begin{equation}
\mathcal{L}_{glb_{s}}=- \frac{1}{n_{s}} \sum_{i=1}^{n_{s}} D_{g}\left(F_2(F_1\left(x_{i}^{s}\right))\right)^{\gamma} \cdot \log \left(1-D_{g}\left(F_2(F_1\left(x_{i}^{s}\right))\right)\right),\label{8}
\end{equation}
\begin{equation}
\mathcal{L}_{glb_{t}}=- \frac{1}{n_{t}} \sum_{i=1}^{n_{t}}\left(1-D_{g}\left(F_2\left(F_1\left(x_{i}^{t}\right)\right)\right)\right)^{\gamma} \cdot \log \left(D_{g}\left(F_2\left(F_1\left(x_{i}^{t}\right)\right)\right)\right),\label{7}
\end{equation}
\begin{equation}
\mathcal{L}_{glb}=\frac{1}{2}\left(L_{glb_{s}}+L_{glb_{t}}\right),\label{9}
\end{equation}
where $n_{s}$ and $n_{t}$ represent the number of source and target images, $x^{s}$ and $x^{t}$ are the target-like and target images, and $F_{1}$ and $F_{2}$ denotes the first seven convolutional layers of the backbone VGG16 and the rest convolutional layers. The detailed structure of global domain classifier $D_g$ is shown in Fig.\ref{fig:d}(a).

Similar with the adversarial training in global alignment, the local domain classifier $D_l$ and shallow layers of the backbone are connected by the GRL. The loss function of local alignment can be written as:
\begin{equation}
\mathcal{L}_{loc_{s}}=\frac{1}{n_{s} H W} \sum_{i=1}^{n_{s}} \sum_{w=1}^{W} \sum_{h=1}^{H} D_{l}\left(F_{1}\left(x_{i}^{s}\right)\right)_{w h}^{2},\label{10}
\end{equation}
\begin{equation}
\mathcal{L}_{loc_{t}}=\frac{1}{n_{t} H W} \sum_{i=1}^{n_{t}} \sum_{w=1}^{W} \sum_{h=1}^{H}\left(1-D_{l}\left(F_{1}\left(x_{i}^{t}\right)\right)_{w h}\right)^{2},\label{11}
\end{equation}
\begin{equation}
\mathcal{L}_{loc}=\frac{1}{2}\left(\mathcal{L}_{loc_{s}}+\mathcal{L}_{loc_{t}}\right),\label{12}
\end{equation}
where $D_{l}\left(F_{1}\left(x_{i}\right)\right)_{w h}$ represents the output of the local domain classifier $D_l$ in each location. The detailed structure of local domain classifier $D_l$ is shown in Fig.\ref{fig:d}(b).

To achieve better adaptation, we regularize the domain discriminator. Previous work has shown that it is effective for stabilizing the adversarial training by regularizing the domain classifier with the segmentation loss in domain adaptive segmentation~\cite{sankaranarayanan2018learning}.Similar with this approach, we regularize the domain discriminator with the detection loss. Formally, we extract the different levels of context vectors $v_1$ and $v_2$ from the middle layers of the domain classifiers $D_l$ snd $D_g$ respectively. Each context vector has 128 dimensions. Then, we concatenate the vectors to obtain the aggregated context vector and all region-wise features are concatenated with the aggregated context to train the domain classifiers to minimize the detection loss and domain classification loss, as illustrated in Fig.\ref{fig2}(b).
\subsection{Overall Objective}
We denote the loss of Faster RCNN~\cite{ren2015faster} as $\mathcal{L}_{det}$ and the overall loss function $\mathcal{L}_{total}$ can be summarized as:
\begin{equation}
\mathcal{L}_{total}=\mathcal{L}_{det}+\mathcal{L}_{rec}+\lambda(\mathcal{L}_{glb}+\mathcal{L}_{loc}),\label{13}
\end{equation}
where $\lambda$ is the hyper-parameter to balance the detection, reconstruction loss and hierarchical alignment losses.
\section{Experiments}
 \begin{table*}[ht]
\centering
\caption{Results (\%) on the adaptation from Cityscapes to Foggy Cityscapes. ‘No DA’ indicates the model is only trained with the source images and directly tested on the target images without any domain adaptation. The best results are in \textbf{bold}, and the second best results are \underline{underlined}.
}
 \resizebox{\linewidth}{!}{
 \begin{tabular}{l|cccccccc|c}
\hline
Methods     & person & rider & car & truck & bus & train & moto & bicycle & mAP \\ \hline
No DA &23.3  &27.9 &32.8 & 11.4 &23.5& 9.3 &12.2& 25.2&
20.7\\
DA~\cite{chen2018domain}$_{\emph{CVPR'2018}}$&25.0 &31.0 & 40.5& 22.1 & 35.3& 20.2 &20.0 &  27.1 &  27.6   \\
DivMatch~\cite{kim2019diversify}$_{\emph{CVPR'2019}}$& 30.8 &40.5 & 44.3& 27.2 & 38.4&  34.5 &28.4  & 32.2 &34.6\\
SWDA~\cite{saito2019strong}$_{\emph{CVPR'2019}}$& 29.9 &42.3  &43.5& 24.5& 36.2& 32.6 &30.0&  35.3 &34.3\\
HTCN~\cite{Chen_2020_CVPR}$_{\emph{CVPR'2020}}$& 33.2& \textbf{47.5} &47.9 &31.6& 47.4& \textbf{40.9}& 32.3& 37.1& 39.8\\
CDN~\cite{su2020adapting}$_{\emph{ECCV'2020}}$&35.8&45.7 & 50.9 &  30.1 &42.5 &29.8 & 30.8 & 36.5 &36.6\\
ATF~\cite{he2020domain}$_{\emph{ECCV'2020}}$&34.6& \underline{47.0} & 50.0& 23.7& 43.3& 38.7& 33.4 &\underline{38.8}& 38.7\\
Progressive~\cite{hsu2020progressive}$_{\emph{WACV'2020}}$ & 36.0 &45.5&  \textbf{54.4} & 24.3&  44.1  &25.8  &29.1  & 35.9 &36.9\\
VDD~\cite{wu2021vector}$_{\emph{ICCV'2021}}$ & 33.4& 44.0& 51.7& \textbf{33.9}& \textbf{52.0} &34.7 &34.2 &36.8& \underline{40.0}\\
CDTD~\cite{shen2021cdtd}$_{\emph{IJCV'2021}}$ &31.6& 
44.0&  44.8 & 30.4 & 41.8 & 40.7&  \underline{33.6} & 36.2&  37.9\\
RPA~\cite{zhang2021rpn}$_{\emph{CVPR'2021}}$ & 33.4& 44.3 &50.1& 29.9& 44.8& \underline{39.1}& 29.9& 36.3& 38.5\\
DDF~\cite{liu2022decompose}$_{\emph{TMM'2022}}$& \textbf{37.2}& 46.3& 51.9 &24.7& 43.9& 34.2 &33.5 &\textbf{40.8} &39.1\\ \hline
FIT-DA (Ours)        &   \underline{36.6}     &   45.8   &   \underline{52.2}  &   \underline{32.2}    & \underline{48.1}   &  34.6    &    \textbf{34.7}  &   37.2     &  \textbf{40.2}   \\ \hline
\end{tabular}}
\label{tab:city}
\end{table*}
\subsection{Datasets}
 We extensively evaluate our approach on three challenging domain adaptive object detection tasks with distinct domain shifts, including adaptation under different weather (Cityscapes~\cite{cordts2016cityscapes} $\rightarrow$ Foggy Cityscapes~\cite{sakaridis2018semantic}), adaptation from the synthetic to the real scene (Sim10K~\cite{johnson2017driving} $\rightarrow$ Cityscapes) and adaptation under different cameras (KITTI~\cite{geiger2013vision} $\rightarrow$ Cityscapes).  Cityscapes~\cite{cordts2016cityscapes} is a dataset of urban street scenes with 8 categories captured with on-board cameras, which has 2975 training images and 500 validating images. Foggy Cityscapes~\cite{sakaridis2018semantic} is the synthetic foggy version of Cityscapes. Sim10K~\cite{johnson2017driving} is a virtual dataset including 10000 images generated by the Grand Theft Auto gaming engine. KITTI~\cite{geiger2013vision} is an autonomous driving dataset that has 7481 images, which is captured by a standard station wagon with two high-resolution video cameras. In the test, we use mean average precision (mAP) metrics for evaluation.
\subsection{Implementation Details}
Our detector is original Faster R-CNN~\cite{ren2015faster} without extra modules. We adopt VGG-16~\cite{simonyan2014very} pre-trained on ImageNet~\cite{deng2009imagenet} as our backbone. In our experiments, the shorter side of the image is resized to 600. Each batch is composed of one source image and one target image. The networks are trained with a learning rate of 0.001 for 50K iterations, then with a learning rate of 0.0001 for 20K more iterations. We use a momentum of 0.9 and a weight decay of 0.0005. $N$ is 64 in Eq.(\ref{1}). For Sim10K $\rightarrow$ Cityscapes, we set $\lambda=0.1$ in Eq.(\ref{13}). For the rest two tasks, we set $\lambda=1$. Our method is implemented with PyTorch.
\subsection{Comparison Experiments }
\textbf{Adaptation under Different Weather. }Table \ref{tab:city} shows the performance of our method on Cityscapes $\rightarrow$ Foggy Cityscapes. We can see that our method alleviates the domain gap across different weather conditions and outperforms all competitors in Table \ref{tab:city}. Compared with GANs-based methods: DivMatch~\cite{kim2019diversify}, Progressive~\cite{hsu2020progressive} and CDTD~\cite{shen2021cdtd}, our method improves the result by +5.6\%, +3.3\% and +2.3\% in mAP, which demonstrates the advantage of the proposed Frequency-based Image Translation for domain adaptive object detection.

\begin{minipage}{\textwidth}
 \begin{minipage}[ht]{0.45\textwidth}
  \centering
     \makeatletter\def\@captype{table}\makeatother\caption{Sim10K to Cityscape.}
       \begin{tabular}{l|c}\hline
Methods     &  mAP   \\ \hline
Source Only &  34.2   \\
DA~\cite{chen2018domain}$_{\emph{CVPR'2018}}$         &  39.0   \\
SWDA~\cite{saito2019strong}$_{\emph{CVPR'2019}}$        & 40.1    \\
HTCN~\cite{Chen_2020_CVPR}$_{\emph{CVPR'2020}}$       &  42.5  \\ 
ATF~\cite{he2020domain}$_{\emph{ECCV'2020}}$&42.8\\
CDTD~\cite{shen2021cdtd}$_{\emph{IJCV'2021}}$&42.6\\
RPA~\cite{zhang2021rpn}$_{\emph{CVPR'2021}}$&\underline{45.7}\\
DDF~\cite{liu2022decompose}$_{\emph{TMM'2022}}$&44.3\\
\hline
FIT-DA (Ours)        &  \textbf{48.6}   \\ \hline
\end{tabular}
\label{tab:gta}
  \end{minipage}
  \begin{minipage}[ht]{0.45\textwidth}
   \centering
        \makeatletter\def\@captype{table}\makeatother\caption{KITTI to Cityscapes.}
         \begin{tabular}{l|c}\hline
Methods     &  mAP  \\ \hline
Source Only &    32.2 \\
DA~\cite{chen2018domain}$_{\emph{CVPR'2018}}$ &  38.5  \\
SWDA~\cite{saito2019strong}$_{\emph{CVPR'2019}}$ &   43.1 \\
CDN~\cite{su2020adapting}$_{\emph{ECCV'2020}}$&  44.9  \\
ATF~\cite{he2020domain}$_{\emph{ECCV'2020}}$&42.1\\
Progressive~\cite{hsu2020progressive}$_{\emph{WACV'2020}}$&  43.9  \\ 
DDF~\cite{liu2022decompose}$_{\emph{TMM'2022}}$&\underline{46.0}\\\hline
FIT-DA (Ours)          &   \textbf{46.3}  \\ \hline
\end{tabular}
\label{tab:KITTI}
   \end{minipage}
\end{minipage}

\begin{table}[ht]
\centering
\caption{Ablation analysis of our method. LA is local feature alignment and GA is global feature alignment. CTV represents the context vector and FIT denotes the frequency-based image translation.}
\begin{tabular}{l|cccc|c|c}\hline
Methods  &\quad LA \quad & GA \quad & CTV \quad & FIT \quad& C $\rightarrow$ F &S $\rightarrow$ C\\ \hline
Source Only&&&&&20.7&34.2  \\\hline
FIT-DA&&&&\checkmark&31.8&40.5  \\
FIT-DA&&\checkmark&\checkmark&\checkmark&38.5&46.4 \\
FIT-DA&\checkmark&&\checkmark&\checkmark&35.7 &42.2  \\
FIT-DA&\checkmark&\checkmark& &\checkmark&37.4 &44.3  \\ 
FIT-DA&\checkmark&\checkmark&\checkmark& &34.5  &40.3  \\ \hline
FIT-DA (Ours)&\checkmark&\checkmark&\checkmark&\checkmark&\textbf{40.2}&\textbf{48.6}  \\ \hline
\end{tabular}
\label{tab:abla}
\end{table}

\noindent
\textbf{Adaptation from the Synthetic to Real Scene. }We evaluate the detection performance on car on Sim10K to Cityscapes benchmark. As we can see the results in Table \ref{tab:gta}, our method has a significant performance boost over other methods, further indicating the effectiveness of our method.

\noindent
\textbf{Adaptation under Different Cameras. }There exists a domain gap between datasets captured through different cameras due to the diversity of hardware devices. We conduct the cross-camera adaptation from KITTI to Cityscapes. The results are presented in Table \ref{tab:KITTI}, and our method has competitive performance among all the comparison methods.

\subsection{Ablation Study}
\textbf{Effectiveness of each component. } We conduct the ablation experiments on Cityscapes $\rightarrow$ Foggy Cityscapes (C $\rightarrow$ F) and Sim10K $\rightarrow$ Cityscapes (S $\rightarrow$ C) to validate the effectiveness of each module in our framework. The results in Table \ref{tab:abla} show that all the modules contribute to the performance improvement (especially FIT module), which indicates the effectiveness of each component in our method. 

\begin{figure*}[ht]
\centering
\includegraphics[width=0.8\textwidth]{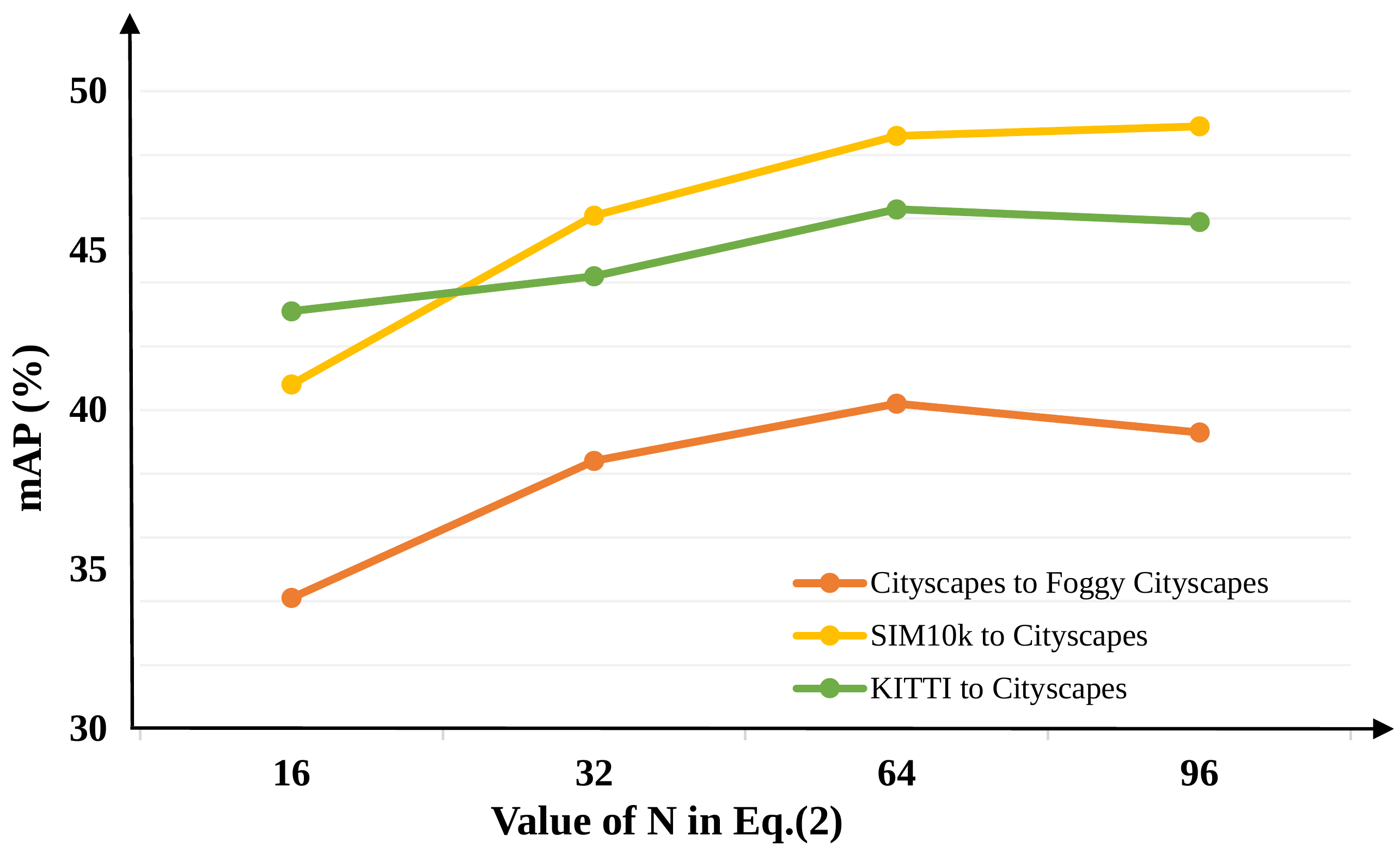}
\caption{Detection performances on the three benchmarks with different values of $N$.
}
\label{fig4}
\end{figure*}

\begin{table}[ht]
\centering
\caption{Performances with different choices of domain-specific frequency components.}
\begin{tabular}{l|c|c|c}\hline
Settings&Choice     & C $\rightarrow$ F &S $\rightarrow$ C\\ \hline
\multirow{4}{*}{Non-learnable}&FC$\left [ 1 \right ]$ &38.3&44.2  \\
&FC$\left [ 1,2 \right ]$ &37.3&45.9 \\
&FC$\left [ 1,2,3 \right ]$ &36.1 &41.8  \\
&FC$\left [ 1,2,3,4 \right ]$ &33.9 &40.0  \\ \hline
Learnable&Frequency Mask&\textbf{40.2}  &\textbf{48.6}  \\ \hline
\end{tabular}
\label{tab:fc}
\end{table}

\noindent
\textbf{Method of choosing domain-specific frequency components. }In FIT framework, Frequency Mask is the core module, which choose the domain-specific frequency component in a learnable way. We compare the adaptation performance of using Frequency Mask with using fixed low-frequency components to determine domain-specific components in Table \ref{tab:fc}. The results suggest the Frequency Mask captures the domain-specific information better. Although low-frequency components largely captures domain-specific information, the distributions of domain-specific components are not completely consistent for images from different domains, making it difficult to capture these components preciously just using fixed low-frequency components. 

\noindent
\textbf{Value of $N$ in Eq.\ref{1}. } $N$ is the number of frequency components after Fourier decomposing. Fig.\ref{fig4} shows the influence on adaptation performance with different $N$. As $N$ is related to the division of frequency bands, which is critical for finding domain-specific frequency components, it affects the quality of translated images. In our experiments, $N=64$ is the best choice considering the performance on the three benchmarks.

\begin{figure}[H]
	\centering
	\begin{minipage}{0.24\linewidth}
        \flushright
		\includegraphics[width=\linewidth]{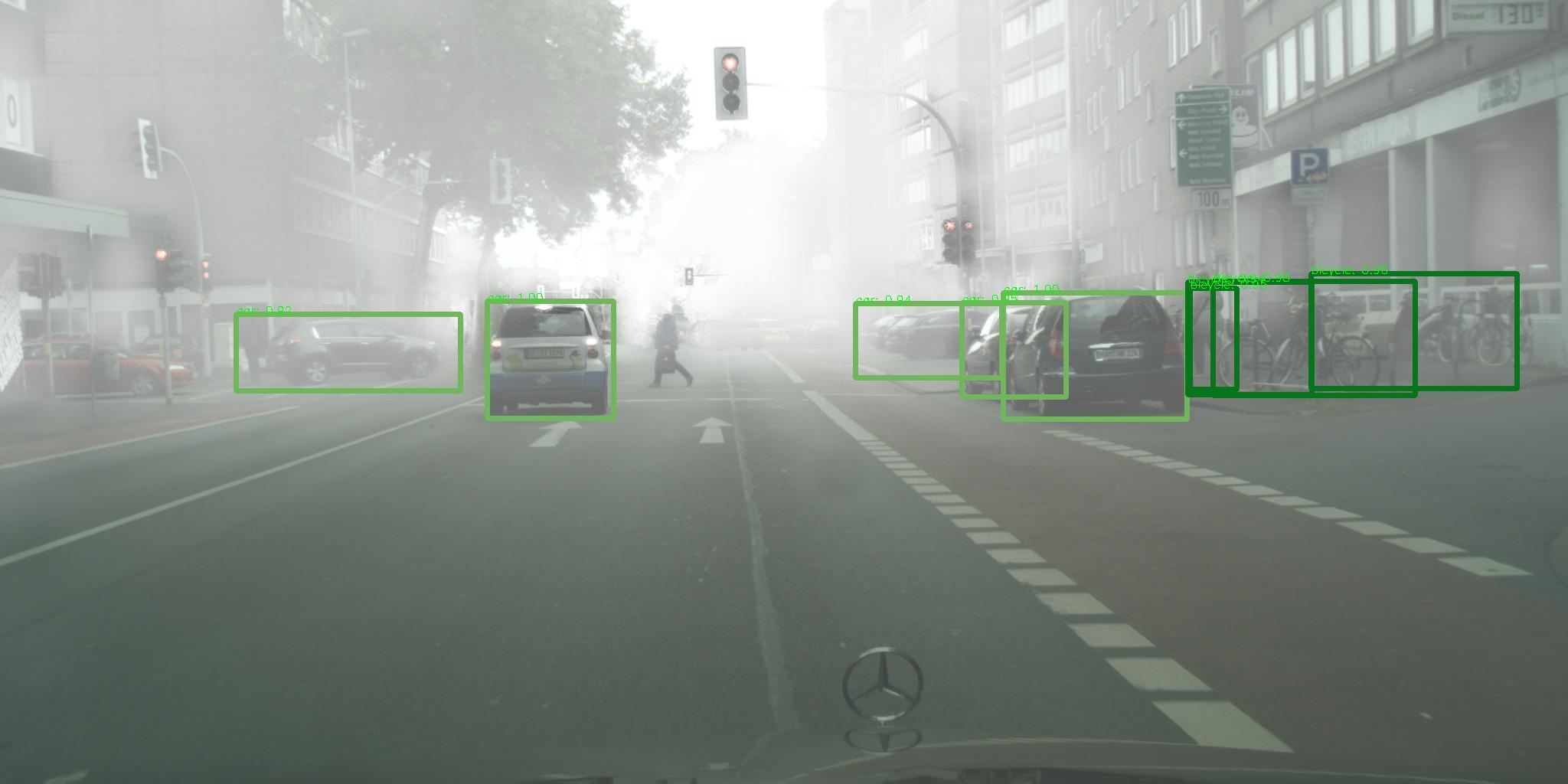}
	\end{minipage}
	\begin{minipage}{0.24\linewidth}
        \flushright
		\includegraphics[width=\linewidth]{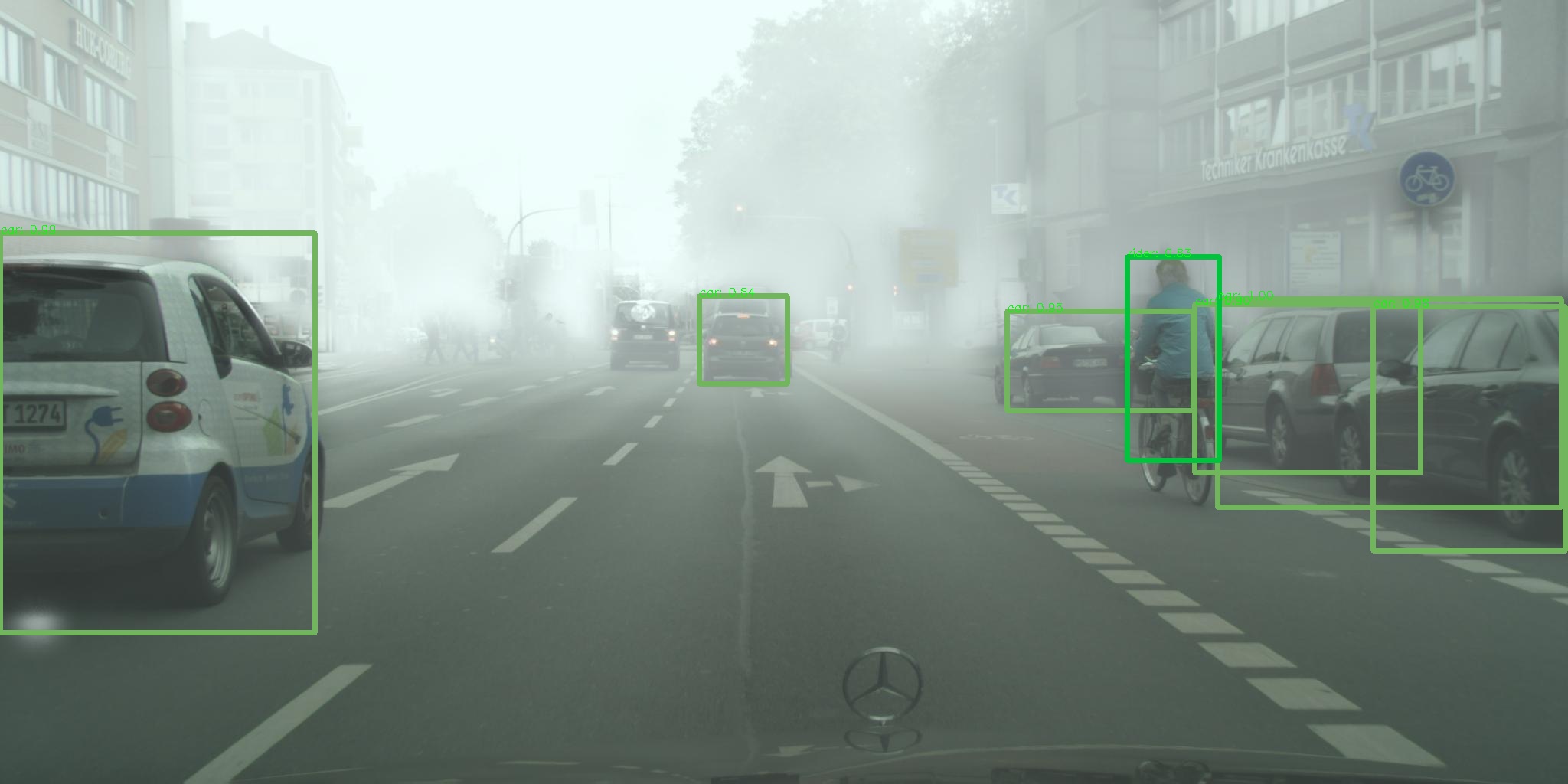}
	\end{minipage}
	\begin{minipage}{0.24\linewidth}
		\flushright
		\includegraphics[width=\linewidth]{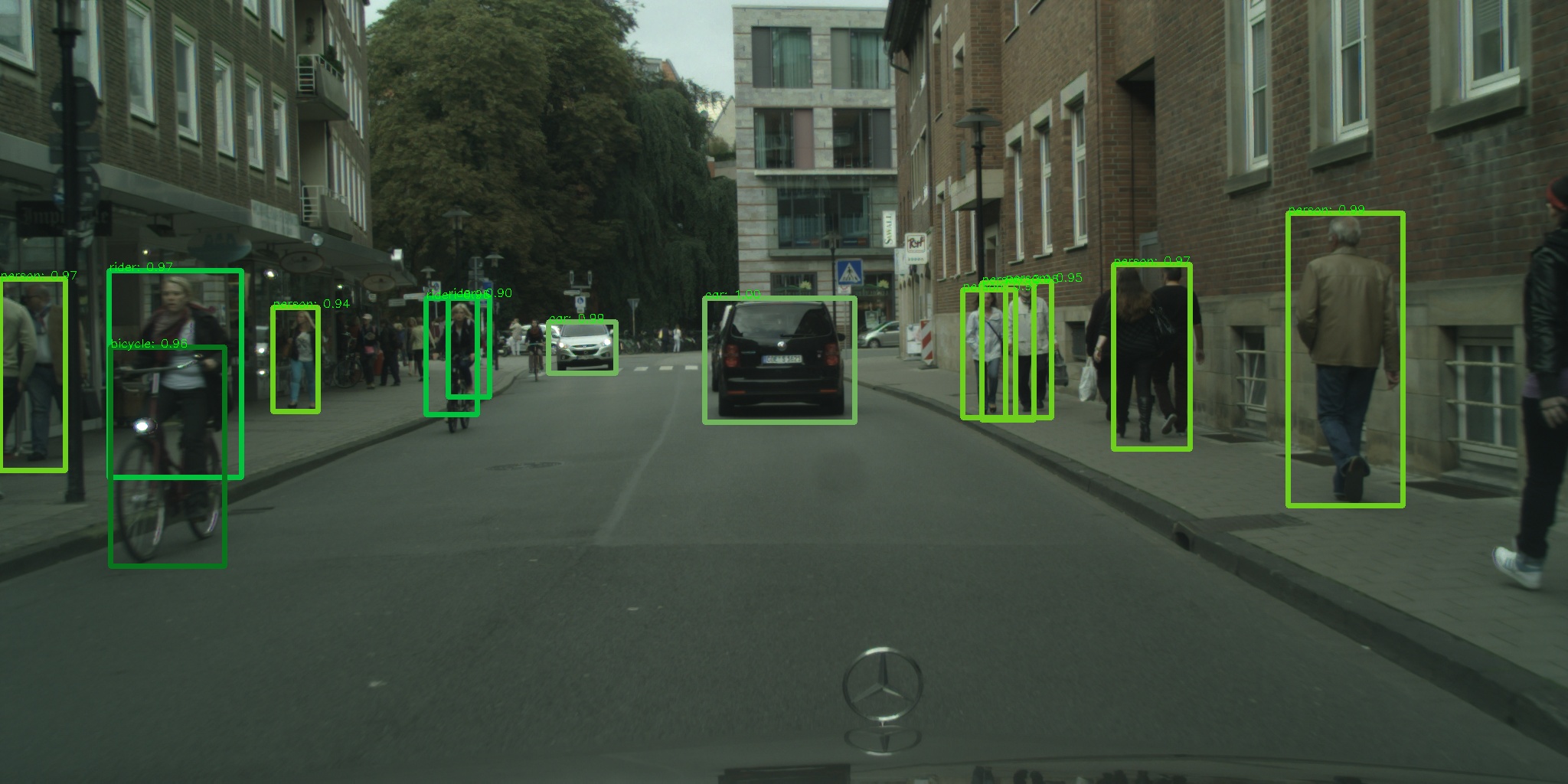}
	\end{minipage}
	\begin{minipage}{0.24\linewidth}
		\flushright
		\includegraphics[width=\linewidth]{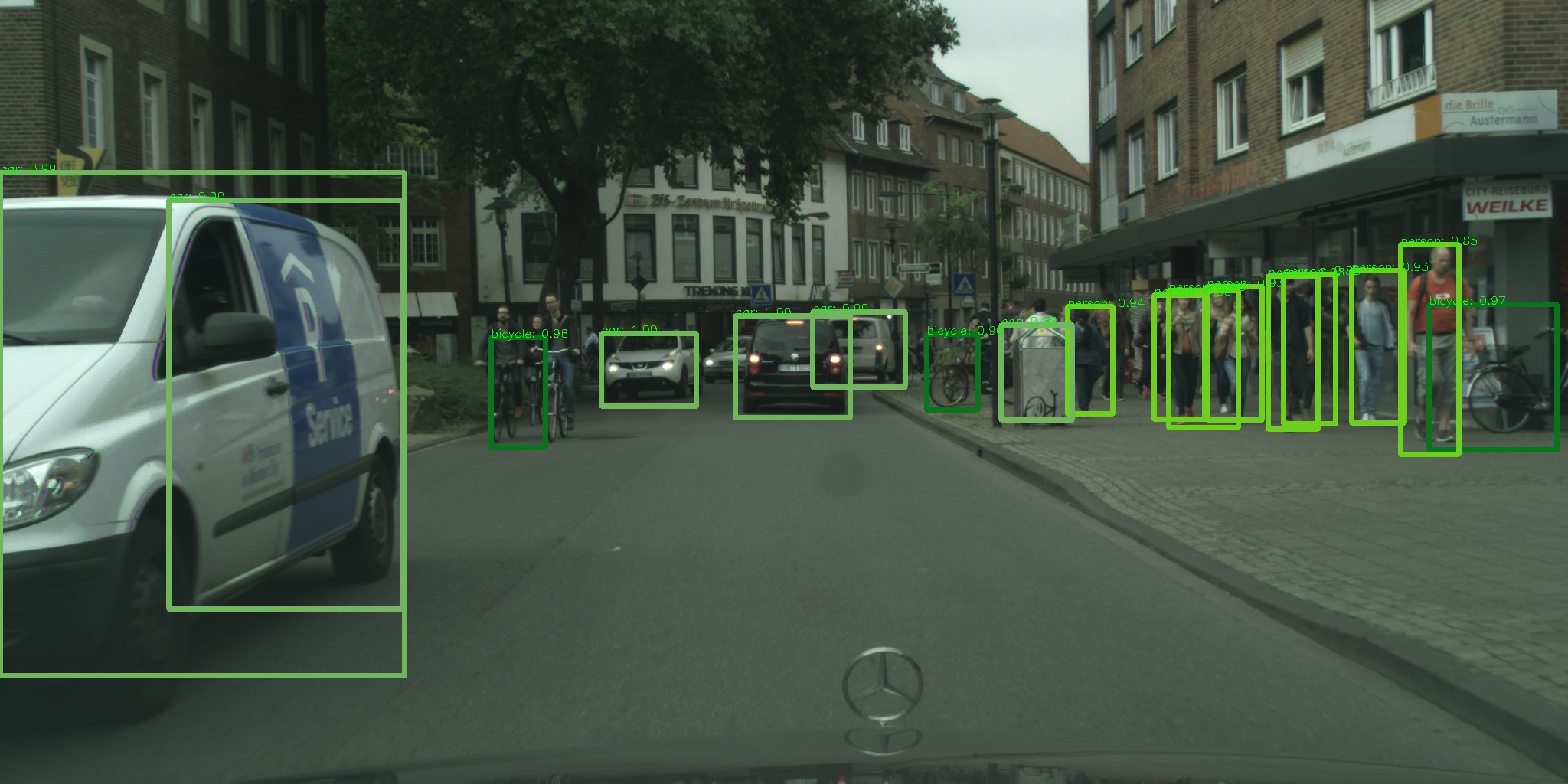}
	\end{minipage}
	
	\begin{minipage}{0.24\linewidth}
        \flushright
		\includegraphics[width=\linewidth]{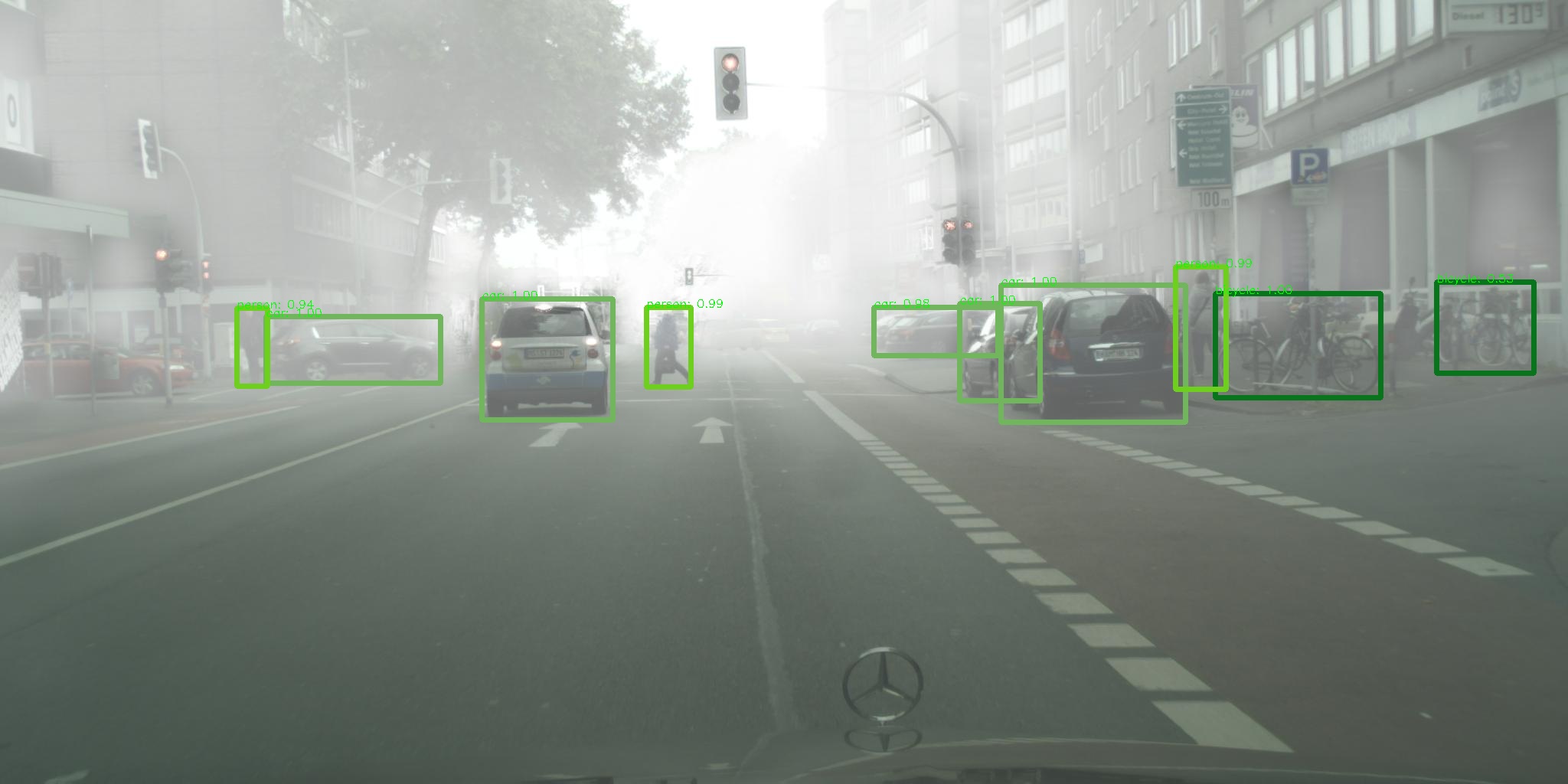}
		\centerline{\quad C $\rightarrow$ F}
	\end{minipage}
	\begin{minipage}{0.24\linewidth}
        \flushright
		\includegraphics[width=\linewidth]{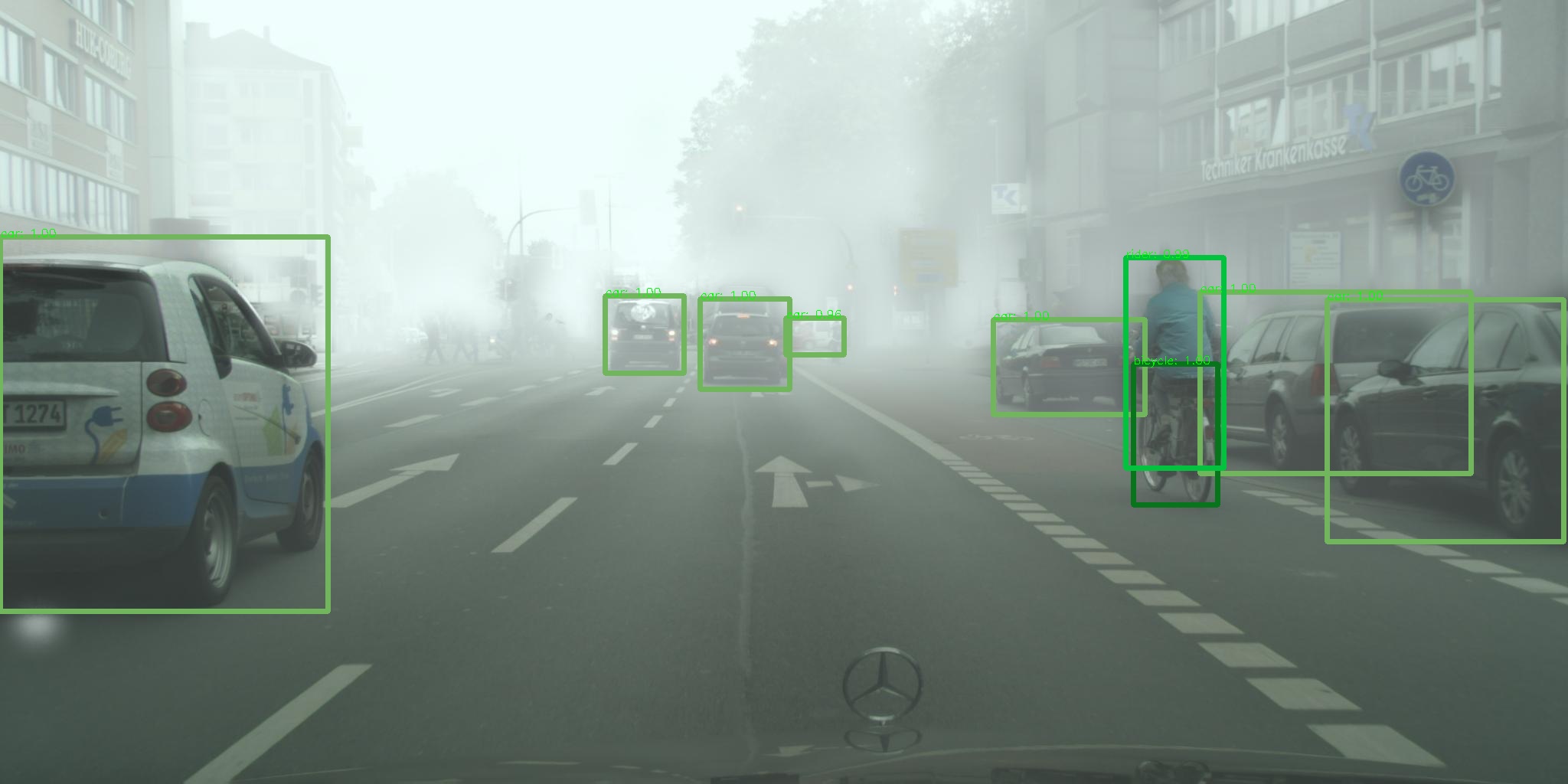}
		\centerline{\quad C $\rightarrow$ F}
	\end{minipage}
	\begin{minipage}{0.24\linewidth}
		\flushright
		\includegraphics[width=\linewidth]{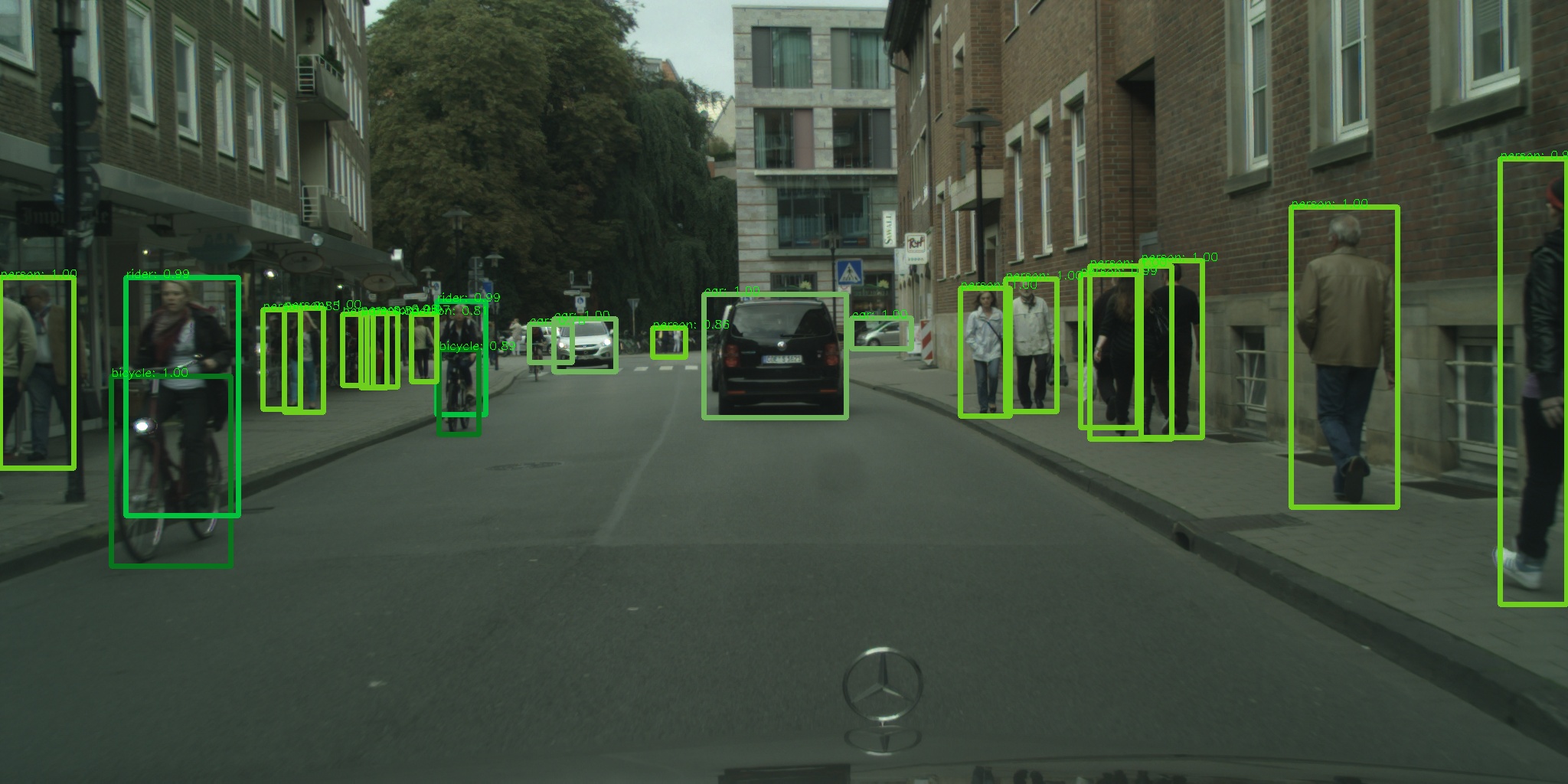}
		\centerline{\quad S $\rightarrow$ C}
	\end{minipage}
	\begin{minipage}{0.24\linewidth}
		\flushright
		\includegraphics[width=\linewidth]{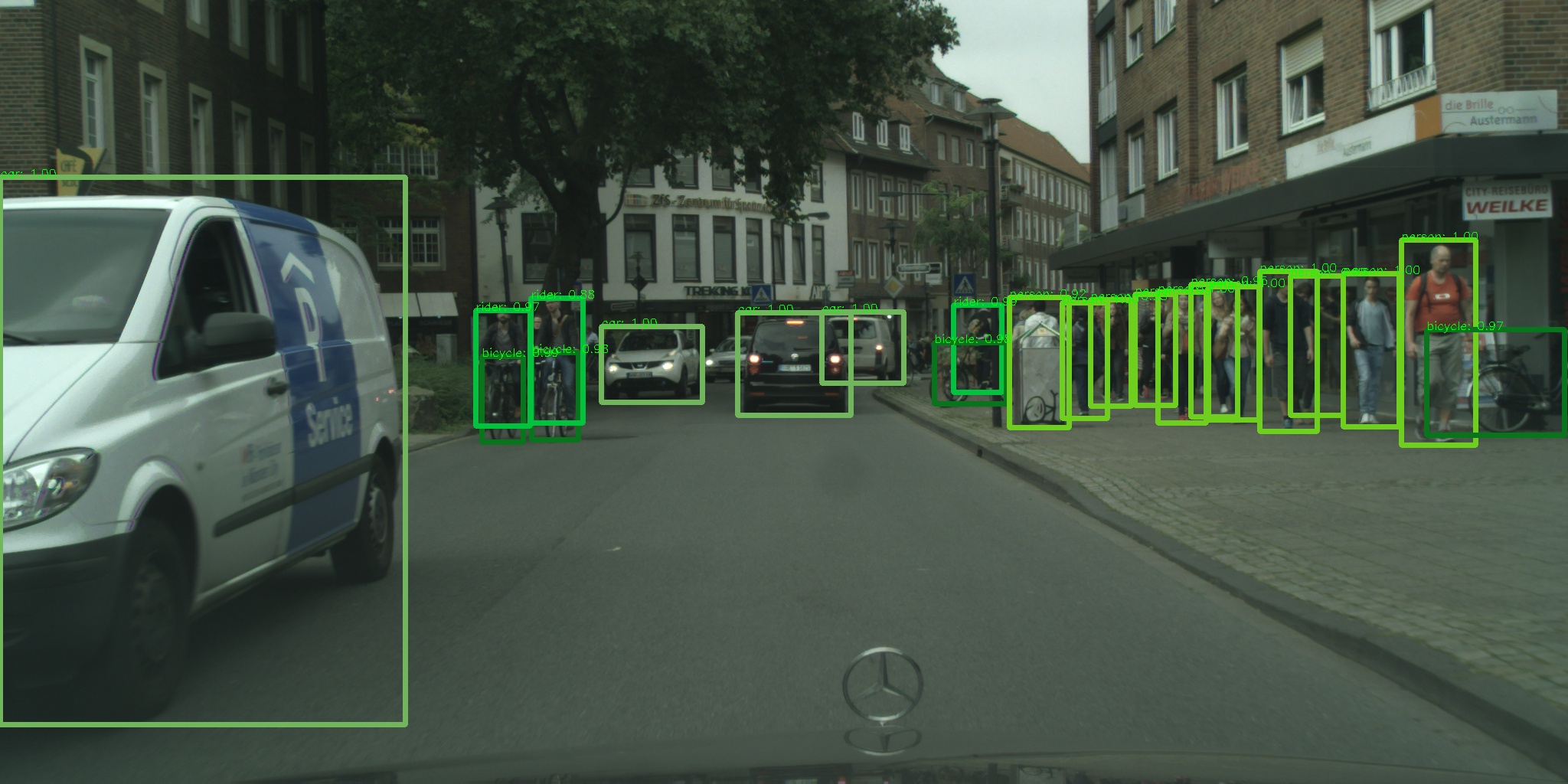}
		\centerline{\quad S $\rightarrow$ C}
	\end{minipage}
\caption{Example results on  Cityscapes to Foggy Cityscapes (C $\rightarrow$ F) and Sim10K to Cityscapes (S $\rightarrow$ C). The fist row is the results of SWDA and the second row is the results of our FIT-DA. The class and score predictions are at the top left corner of the bounding box. Zoom in to visualize the details.} 
\label{fig5}
\end{figure}

\subsection{Visualization}
Fig.\ref{fig5} illustrates some examples of detection results on on Cityscapes to Foggy Cityscapes and Sim10K to Cityscapes. Obviously, our method produces more accurate bounding box predictions and has a stronger ability to detect obscured instances.
\section{CONCLUSION}
In this paper, a novel Frequency-based Image Translation (FIT) method for DAOD is presented to reduce domain shift at the input level. Compared to other image translation methods for DAOD, it is embedded in the detection network and does not need extra time-consuming training. Additionally, we introduce hierarchical adversarial feature learning to further mitigate the domain gap at the feature level. Meanwhile, a joint loss function is designed to optimize the entire network in an end-to-end manner. Extensive experiments on three challenging DAOD benchmarks validate the effectiveness of our method. In the future, we will utilize the frequency information in the feature space to investigate the feature augmentation for DAOD.

\subsubsection{Acknowledgements} This work was supported in part by the National Key Research and Development Plan of China under Grant 2020AAA0108902 and the Strategic Priority Research Program of Chinese Academy of Science under Grant XDB32050100.
%
%
%

\end{document}